# DEMAND: Deep Matrix Approximately Nonlinear Decomposition to Identify Meta, Canonical, and Sub-Spatial Pattern of functional Magnetic Resonance Imaging in the Human Brain

**Wei Zhang**
University of California San Francisco
wei.zhang4@ucsf.edu

**Yu Bao**
James Madison University
bao2yx@jmu.edu

## Abstract

Deep Neural Networks (DNNs) have already become a crucial computational approach to revealing the spatial patterns in the human brain; however, there are three major shortcomings in utilizing DNNs to detect the spatial patterns in functional Magnetic Resonance Signals: 1). It is a fully connected architecture that increases the complexity of network structures that is difficult to optimize and vulnerable to overfitting; 2). The requirement of large training samples results in erasing the individual/minor patterns in feature extraction; 3). The hyperparameters are required to be tuned manually, which is time-consuming.

Therefore, we propose a novel deep nonlinear matrix factorization named Deep Matrix Approximately Nonlinear Decomposition (DEMAND) in this work to take advantage of the shallow linear model, e.g., Sparse Dictionary Learning (SDL) and DNNs. At first, the proposed DEMAND employs a non-fully connected and multilayer-stacked architecture that is easier to be optimized compared with canonical DNNs; furthermore, due to the efficient architecture, training DEMAND can avoid overfitting and enables the recognition of individual/minor features based on a small dataset such as an individual data; finally, a novel rank estimator technique is introduced to tune all hyperparameters of DEMAND automatically.

Moreover, the proposed DEMAND is validated with four other peer methodologies via real functional Magnetic Resonance Imaging data in the human brain. In short, the validation results demonstrate that DEMAND can reveal the reproducible meta, canonical, and sub-spatial features of the human brain more efficiently than other peer methodologies.

## 1  Introduction

Hierarchical functional structures in the human brain [1-4] have been uncovered by multiple techniques of deep matrix factorization, such as Low-to High-Dimensional Independent Components Analysis (DICA) [5], Sparse Deep Dictionary Learning (SDDL) [6], Deep Non-negative Matrix Factorization (DNMF) [7], [8]. Furthermore, with the development of deep learning methods, a variety of deep neural networks (DNNs) have provided an opportunity to reconstruct hierarchical features in the brain, e.g., the Deep Convolutional Auto Encoder (DCAE), Deep Belief

Network (DBN), and Convolutional Neural Network (CNN) [9-15]. In detail, the Restricted Boltzmann Machine (RBM) can extract hierarchical temporal features from functional Magnetic Resonance Imaging (fMRI) and effectively reconstruct functional connectivity (FC) networks with impressive accuracy [16, 17]. In addition, other recent research has found the reasonable hierarchical temporal organization of task-based fMRI time series, each with corresponding task-evoked FCs [9, 10, 16, 17] using DCAE, RBM, and DBN. These machine learning techniques are generally considered deep nonlinear models, i.e., DNNs, constructed with nonlinear activation functions [18].

Specifically, these methodologies that reveal the hierarchical FCs in the brain can be divided into two folds: 1). Deep Linear Model where these methods concentrate on adopting a deep strategy of matrix decomposition, e.g., DNMF, SDDL; 2). Deep Nonlinear Models where these methods concentrate on a fully-connected architecture with activation functions to provide a more substantial perception, e.g., DBN, DCAE. Moreover, the vital advantage of a deep linear model that it is easy to be optimized and does not require large training samples due to the convex optimization function and the non-fully connected architectures; in contrast, the utilization of the nonlinearity and fully connected structures in deep nonlinear models enables the substantial perception but also increases the training difficulty and requires large training samples to avoid overfitting; furthermore, training a deep nonlinear model on a group-wised dataset usually easily vanishes the individual/minor features. Although a variety of nonlinear models, such as DBN, Deep Boltzmann Machine (DBM), and DCAE, have recently shown the effectiveness in the hierarchical spatiotemporal reconstruction of task-evoked fMRI data [8-10, 17], there are still six shortcomings to overcome: 1) large training samples [19-22]; 2) extensive computational resources, e.g., graphics processing units (GPUs) or tensor processing units (TPUs) [10, 11, 13]; 3) manual tuning of all hyperparameters, e.g., the number of layers and size of a dictionary, whereas some parameters such as sparse trade-off and step-length, are not denoted as hyperparameters [8-10]; 4) time-consuming training process [13, 14]; 5) uncertainty of convergence to the global optimum [13, 14, 19, 20, 22]; and 6) "black box" results that are challenging to explain [13, 14, 19].

**Contributions**. We propose Deep Matrix Approximately Nonlinear Decomposition (DEMAND) that aims to benefit from both deep linear and nonlinear models. To be specific,

1). *Efficient Non-Fully Connected Architecture*. Unlike deep linear and nonlinear models, DEMAND employs non-fully connected architectures that reduce the computational complexity compared to other fully connected deep neural networks. The fundamental difference in the structure of DEMAND and that of DNNs is presented in Figure S1 in Appendix A, supplementary material. The non-fully connected architectures enable more efficient training than DNNs and resolve the overfitting issue when the training sample is small, e.g., a single subject. Moreover, DEMAND employs the same activation functions used in DNNs to maintain the substantial perception [13, 19]. Finally, we theoretically analyze the superiority of non-fully connected structures. This theoretical analytics is presented as Theorem 1.1 in Appendix A.

2). *Automatic Hyperparameters Tuning Technique*. To implement the automatic tuning of hyperparameters, e.g., number of layers, size of components, and number of units/neurons in each layer [23, 24], we introduce a rank reduction technique named rank reduction operator (RRO) for DEMAND. Specifically, RRO utilizes the orthogonal decomposition, e.g., *QR* decomposition, to estimate the rank of feature matrices (number of features) via the weighted ratio (WR), the weighted difference (WD), and the weighted correlation (WC). These three techniques can consistently reduce the size of feature matrices at all layersif the estimated number of features is equal to one that indicates the decomposition is finished. Thus, due to the generalization and efficacy of *QR* decomposition, RRO can tune all hyperparameters faster than Singular Value Decomposition (SVD) adopted by Principal Component Analysis (PCA) [23, 24]. The details of RRO implementation can be viewed in Algorithms 2.1, 2.2, 2.3, and 2.4, Appendix B, in Supplementary Material. The theoretical analytics of RRO can be viewed in Theorem 3.4 included in Appendix C, Supplementary Material.

3). *Reduced Accumulative Error via Matrix Backpropagation*. Given that the accumulative error could potentially decrease the reconstruction accuracy, we utilize a technique of matrix backpropagation (MBP) [6, 7, 25] to reduce the accumulative error. The detail of MBP implementation can be found in Algorithm 2.5 in Appendix B, Supplementary Material.

4). *Accurate Approximation to Input Signal Matrix*. We have proved that DEMAND has comparable approximation accuracy even with DNNs. The conclusion and proof details are presented in Theorem 3.1 in Appendix C, Supplementary Material.

5). *Fast Convergence.* Given Adam as an efficient optimizer to update all variables of DEMAND [26], our theoretical analysis demonstrates that DEMAND can maintain the convergence speed as Adam itself, according to Theorem 3.5, Appendix C, Supplementary Material. Moreover, given the continuously increasing high dimensionality of the dataset, we also discuss the convergence of DEMAND on an infinite dimensionality space, where all proofs can be found in Collaroy 3.3 in Appendix C, Supplementary Material.

***Related Works and Methodological Validation***. In short, the proposed DEMAND is validated on real resting-state fMRI signals and compared with other peer four methods. The validation results demonstrate that DEMAND can detect reproducible meta, canonical, and sub-spatial features in the human brain than deep linear/nonlinear models and is easier to be optimized than deep nonlinear models that is DEMAND will be quickly converged with a more accurate training loss. Furthermore, some deeper spatial features derived from DEMAND can be treated as meta-spatial features that usually contain several functional regions of the shallow FC networks. In contrast, the other minor features are denoted as sub-FC and have been not fully reported in identifying of sub-FCs via current methodologies.

## 2  Method

This section provides the details of the five vital contributions mentioned above of DEMAND: the efficient model architecture, the automatic hyperparameters tuning, the matrix backpropagation, the accurate approximation, and the convergence.

### 2.1  Deep Linear Matrix Approximation for Reconstruction

The proposed DEMAND employs an efficient non-fully connected architecture with the nonlinear activation function used in DNNs. In detail, the optimization function governing DEMAND is:

$$min_{S_i \in \mathbb{R}^{m \times n}} \bigcup_{i=1}^{k} \|S_i\|_1$$
$$s.t. \forall k \in [2, M], (\prod_{i=1}^{k} X_i) \cdot Y_k + S_k = I \quad (1)$$
$$X_i Y_i \leftarrow \mathcal{R} \cdot (\mathcal{N}_{i-1} \cdot Y_{i-1}), \forall\, 2 \leq i \leq k$$

where $\{X_i\}_{i=1}^{k}$ represents the hierarchical weight matrices or mixing matrices, e.g., $X_i$ indicates the weight matrix of the $i^{th}$ layer. In addition, $M$ is the total number of layers. Similarly, $\{Y_i\}_{i=1}^{k}$ represents the hierarchical FCs, i.e., the spatial features; for instance, $Y_i$ indicates the spatial features of the $i^{th}$ layer. Furthermore, $\{S_i\}_{i=1}^{k}$ is a set of matrices of the background FCs, which are usually treated as noisy patterns, due to their sparsity. Moreover, $\mathcal{R}$, representing the RRO operator, reduces the high dimensionality and automatically estimate the hyperparameters; meanwhile $\mathcal{N}_{i-1}$ represents the nonlinear activation function of i-1 layer, e.g., Sigmoid [14, 20]. And the original input data $I$ is supposed to be decomposed as $(\prod_{i=1}^{M} X_i) \cdot Y_M + S_M$, if we assume the total number of layers is $M$.

In Eq. (1), our fundamental assumption is the spatial features $Y_{i-1}$ at the previous layer can be continuously decomposed as a product of deeper weight matrix $X_i$ and spatial features $Y_i$. And this optimization function of DEMAND consists of more variables than conventional deep linear models, such as DICA and SDDL.

Before optimizing Eq. (1), we need to convert Eq. (1) into an augmented Lagrangian function. Considering the $k^{th}$ layer, we have:

$$\mathcal{L}_\lambda(\prod_{i=1}^{k} X_i, Y_k, S_k, \mathcal{N}_k) \stackrel{\text{def}}{=} \frac{\lambda}{2} \left\| (\prod_{i=1}^{k} X_i) \cdot [\mathcal{R} \cdot (\mathcal{N}_k \cdot Y_k)] - I \right\|_F^2 + \frac{1}{\lambda} \|S_k\|_1 \quad (2)$$

The sparse trade-off of $\bigcup_{i=1}^{k} \|S_i\|_1$ controlling the sparsity levels of background components is determined by $\frac{1}{\lambda}$ that can also be estimated using Rose Algorithm [27]. Naturally, it is easier to employ alternative strategies and shrinkage methods [25, 26] to minimize Eq. (2). Due to the efficiency of the Adam optimizer [26], we adopt Adam to update all the variables in DEMAND.

Furthermore, the $\ell_1$ norm of $S_k$ shown in Eq. (1) can be solved directly using the shrinkage method [28].

Denote Adam [26] as an operator $\mathcal{A}$. The alternative iterative format of Adam to update all the variables in Eq. (2) can be organized as follows:

$$X_k^{it+1} \leftarrow \mathcal{A} \cdot (X_k^{it}) \tag{3-1}$$

$$Y_k^{it+1} \leftarrow \mathcal{A}(\mathcal{N}_k \cdot Y_k^{it}) \tag{3-2}$$

$$S_k^{it+1} \leftarrow Shrinkage\ [(\prod_{i=1}^{k} X_k^{it+1}) \cdot \mathcal{R} \cdot (\mathcal{N}_k \cdot Y_k^{it}) - I] \tag{3-3}$$

In detail, in Eq. (3-1), $X_k^{it}$ (the current iteration is represented as *it*), is updated by the Adam; meanwhile, $Y_k^{it}$ is treated as a constant; similarly, in Eq. (3-2), we only update a single variable $Y_k^{it}$ independently. Finally, Eq. (3) demonstrates the shrinkage and minimization of the background component or the noise feature $S_k^{it+1}$.

## 2.2 Rank Reduction Operator for Automatic Tuning Hyperparameters

To reduce the high dimensionality and tune the hyperparameters without a manual design, we propose a technique named RRO. In short, since RRO can continuously estimate the rank of the current feature matrix until the rank of the feature matrix is reduced to 1, using RRO, DEMAND enables the automatic hyperparameters tuning. In detail, rank estimator RRO employs rank-revealing by constantly using orthogonal decomposition via *QR* factorization [23, 24]. The advantage of *QR* factorization is that it is faster than Singular Value Decomposition (SVD) and needs fewer requirements of the input matrix. For instance, SVD cannot easily decompose the sparse matrix [23, 24]. The mathematical definition of RRO is shown below:

$$\mathcal{R}\begin{bmatrix}a_1\\a_2\\\vdots\\a_{n-1}\\a_n\end{bmatrix} = \begin{bmatrix}a_1^{(1)}\\a_2^{(1)}\\\vdots\\a_{n-2}^{(1)}\\a_{n-1}^{(1)}\end{bmatrix} \mathcal{R}^k\begin{bmatrix}a_1\\a_2\\\vdots\\a_{n-1}\\a_n\end{bmatrix} = \begin{bmatrix}a_1^{(1)}\\a_2^{(1)}\\\vdots\\a_{n-k-1}^{(1)}\\a_{n-k}^{(1)}\end{bmatrix} = [\hat{a}] \tag{4}$$

where $\mathcal{R}$ denotes the RRO operator. Also, by defining $\{a_i\}_{i=1}^n$ as a series of vectors and $\hat{a}$ as a single vector, we have $rank([\hat{a}]) < rank(\mathcal{R}^k \cdot [a_1, a_2, \cdots, a_n])$; if $k$ is large enough, e.g., $\exists N \in \mathbb{N},\ k > N$, then we have $rank([\hat{a}]) = 1$, and $k$ is equivalent to the total number of layers, due to the feature matrix is equal to a single vector after performing RRO for $k$ times. Importantly, we prove $\mathcal{R}: \mathbb{R}^{S \times T} \to \mathbb{R}^{S \times T}, \|\mathcal{R}\| < \infty$, meaning that $\mathcal{R}$ is a bounded operator in Theorem 3.4, Appendix C, Supplementary Material.

Furthermore, denote $r^*$ as the initially estimated rank and $r$ as the optimal rank estimation of the input signal matrix $S$. If $r^* \geq r$ holds, the diagonal line of the upper-triangular matrix of signal matrix $S$ can be detected after *QR* factorization and previously introduced three techniques such as *WR*, *WD*, and *WC* can be utilized to estimate the rank. In detail, at first, the diagonal matrix $R$ is non-increasing in magnitude [23, 24]; furthermore, along the main diagonal of matrix $R$, *WR*, *WD*, and *WC*, are used to estimate the maximum rank in Eqs (5)-(7); next, $S$ will be replaced by $Y_k\ k = 1,2,\cdots,M$, iteratively. Thus, using *QR* factorization, the rank-reducing technique will eventually generate a reasonable solution [23, 24].

Denote $d \in \mathbb{R}^{1 \times r}$ and $r \in \mathbb{R}^{r-1}$, then *WR* can be calculated by Eq. (5):

$$\begin{aligned}d_i &\leftarrow |R_{ii}|\\wr_i &\leftarrow \frac{d_i}{d_{i+1}}\end{aligned} \tag{5}$$

where $R_{ii}$ represents a diagonal element of matrix $R$ derived by *QR* decomposition and $wr_i$ represents a single value of *WR*. The value of each *WR* is calculated by the ratio of the current element of diagonal and the following element.

Similarly, *WD* is calculated as:

$$wd_i \leftarrow \frac{|d_i - d_{i-1}|}{\sum_{k=1}^{i-1} d_k} \quad (6)$$

In Eq. (6), *WD* is defined as the absolute difference between the current diagonal element and the previous one divided by the cumulative sum of all the previous diagonal elements.

Eq. (7) describes the proposed *WC* as follows:

$$wc_i \leftarrow \frac{|corr(d_{i-2}, d_{i-1}) - corr(d_{i-1}, d_i)|}{\sum_{k=1}^{i+1} d_k} \quad (7)$$

Since *WD*, *WR*, and *WC* are the cumulative difference, ratio, and correlation of two adjacent components, respectively, the matrix dimension can be reduced by one at least after rank estimation. Thus, the RRO iteratively determines the maximum value position from *WR*, *WD*, and *WC* to conduct the estimated rank. The details of pseudocodes to implement *WR*, *WD*, and *WC* can be viewed in Algorithm 2.2, 2.3, and 2.4 in Appendix B, Supplementary Material.

## 2.3 Matrix Backpropagation

To reduce the potential accumulative errors caused by a large number of layers, we introduce a matrix backpropagation technique to increase the reconstruction accuracy, e.g., $\left\|\prod_{i=1}^{k} X_i \left[\mathcal{R} \cdot (\mathcal{N}_k \cdot Y_k)\right] - I\right\|_F^2$. Specifically, the accumulative error can be reduced by the derivative of nonlinear activation functions [6, 7, 25].

$$K \leftarrow (\prod_{k=1}^{M-1} X_k^{it})^T \cdot I \quad (8\text{-}1)$$

$$P_k^{it} \leftarrow (X_k^{it})^T X_k^{it} \cdot \prod_{k=1}^{M-1} \max(X_k) \quad (8\text{-}2)$$

In Eqs. (8-1) and (8-2), there are two important variables calculated for the following backpropagation techniques shown as Eqs. (9) and (10) [6, 25]. The following equations denote the two variables $c_k^{it}$ and $V_k^{it}$ to perform backpropagation [6, 7] using the derivative of the inverse activation function $\frac{d\mathcal{N}^{-1}(x)}{dx}$:

$$c_k^{it} \leftarrow \max(X_k) \cdot \frac{d\mathcal{N}^{-1}(x)}{dx}, x = X_k^{it} Y_k^{it} \quad (9\text{-}1)$$

$$V_k^{it} \leftarrow (X_k^{it})^T \cdot (P_k^{it} \cdot \mathcal{N}^{-1}(x) - K) \odot c_k^{it}, x = X_k^{it} Y_k^{it} \quad (9\text{-}2)$$

Then, the following equations show the process of updating the weight matrix $X_k$ and the feature matrix $Y_k$ for the $k^{th}$ layer. In addition, $E$ is a constant value initialized as 0.01 [6, 7, 25].

$$Y_k^{it+1} \leftarrow Y_k^{it} - \frac{E}{2^{it}}(V_k^{it}) \quad (10\text{-}1)$$

$$X_k^{it+1} \leftarrow X_k^{it+1} - \frac{E}{2^{it}}(V_k^{it}) \quad (10\text{-}2)$$

## 2.4 Accurate Approximation, Efficient Architecture, and Fast Convergence of DEMAND

This section theoretically explains the efficacy of non-fully connected architectures and analyzes the approximation and convergence of DEMAND. Since DEMAND is treated as a composition of linear and nonlinear functions, the following theorem demonstrates that DEMAND can approximate any real function that is almost-everywhere infinite [29], with very high accuracy. Nevertheless, only infinite layers can guarantee that the reconstruction accuracy is comparable to Deep Linear Models. The proof of Theorem 1.1 can be viewed in Appendix A, Supplementary Material.

***Theorem 1.1*** **(Composition of Non-Smooth Activation Functions)** Given a non-smoothed activation function $f_i$ with a single non-smooth point, denoted on $[a,b] \subseteq \mathbb{R}^1$, $f_i \in Lip1([a,b] \setminus \{U(x_i, \delta)\})$ $i \in \mathbb{N}$. $U(x_i, \delta)$ is an open cubic with the center $x_i$ and radius $\delta > 0$. The composition of $f_i$ and $f_j$ is denoted as $f_{j,i} \stackrel{\text{def}}{=} f_j(f_i(x))$, and multiple composition $\mathcal{F} \stackrel{\text{def}}{=} f_{\cdots,k,\cdots j.i} \in Lip1([a,b] \setminus [c,d])$, when $k \to \infty$, $[c,d] \supseteq \cup_{i=1}^{k} U(x_i, \delta)$ and $m([c,d]) \neq 0$; moreover, given $t \to \infty$, the summation of $\sum_{i=1}^{t} \mathcal{F}_i$ leads to $\sum_{i=1}^{t} \mathcal{F}_i \notin Lip1([a,b] \setminus [c',d'])$, $[c',d'] \supseteq [c,d]$, and $m([c',d']) \neq 0$. In addition, $m(\cdot)$ represents the Lebesgue measure.

Theorem 1.1 indicates the infinite composition and summation of activation function, such as fully-connected and very deep neural network architecture, with a single non-smooth point that can become a function that is non-smooth in an interval. Furthermore, this theorem demonstrates that the non-fully connected architectures can be more easily optimized.

***Theorem 3.1*** **(Accurate Approximation of DEMAND)** Given a real function $f: \mathbb{R}^{S \times T} \to \mathbb{R}^{S \times T} \cup \{\pm\infty\}$ and $m(\{X \in \mathbb{R}^{S \times T}: f(X) = \pm\infty\}) = 0$ where $m(\cdot)$ represents the Lebesgue measure [29]. Given $X \in \mathbb{R}^{S \times T}$ and a serie of linear composition of nonlinear activation functions denoted as $\{\mathcal{N}_k(X)\}_{k=1}^{N}$, we have: if $\mathcal{N}$ denotes a smooth activation function, and $\forall \varepsilon > 0$, $N < \infty$, $\|\{\mathcal{N}_k(X)\}_{k=1}^{N} - f(X)\| \leq \varepsilon$.

Theorem 3.1 demonstrates that DEMAND enables a very accurate approximation as $\forall \varepsilon > 0$, $\|\{\mathcal{N}_k(X)\}_{k=1}^{N} - f(X)\| \leq \varepsilon$ to the original input $f(X)$ even is almost-everywhere finite, e.g., $f(X) = \pm\infty, m(X) = 0$. Moreover, to reach an accurate approximation, DEMAND only requires limited layers, such as $N < \infty$, and $\{\mathcal{N}_k(X)\}_{k=1}^{N}$.

***Definition 3.1*** **(DEMAND Operator)** Denote the Random Initialization Operator as $\mathcal{R}: \mathbb{R}^{m \times n} \to \mathbb{R}^{m \times n}$, the Sparse Operator as $\mathcal{S}: \mathbb{R}^{m \times n} \to \mathbb{R}^{m \times n}$ and the Adam operator as $\mathcal{A}: \mathbb{R}^{m \times n} \to \mathbb{R}^{m \times n}$. Their norms can be represented as $\frac{1}{r} \stackrel{\text{def}}{=} \|\mathcal{R}\|$, $\frac{1}{s} \stackrel{\text{def}}{=} \|\mathcal{S}\|$, and $\frac{1}{a} \stackrel{\text{def}}{=} \|\mathcal{A}\|$; in Lemma 4.1 to 4.4 in Appendix D, Supplementary Material, we have: $0 < \frac{1}{s} < 1$ and $0 < \frac{1}{r}, \frac{1}{a} < \infty$, but $0 < \frac{1}{a^k} < 1, k > N$.

***Theorem 3.2*** **(Convergence of DEMAND in Finite Dimensionality Space)** Denote Adam as an operator $\mathcal{A}$ in a finite dimensionality space. Due to the convergence of Adam with a velocity of $\mathcal{O}(\frac{1}{\sqrt{T}})$, the convergence of DEMAND is guaranteed to be the same as Adam.

Theorem 3.2 shows that DEMAND can converge as fast as Adam due to the prerequisite of finite dimensionality space. We further prove the convergence of DEMAND in an infinite dimensionality space in Collaroy 3.1.

***Collaroy 3.1*** **(Convergence of DEMAND in Infinite Dimensionality Space)** Given the infinite dimensionality space [30, 31], denote DEMAND as an operator $\mathcal{D}: \mathbb{R}^{\infty \times \infty} \to \mathbb{R}^{\infty \times \infty}$. If assume $\mathcal{D}$ is an infinite dimensional matrix operator with each element represented as $d_{i,j}$ is an element of $\mathcal{D}$ and $r_{i,j} \in \mathbb{R}^{\infty \times \infty}$, $i,j \to \infty$, respectively. And $\mathcal{D}$ can converge to a fixed point, if and only if $d_{i,j}^k \cdot r_{i,j}$ is $\mathcal{O}\left(\frac{1}{n^p}\right), n \in \mathbb{N}, p > 1, p \in \mathbb{R}$.

***Theorem 4.1*** **(Convergence of DEMAND using Alternative Update)** Given $\{\mathcal{F}_{i,j}\}_{i,j=1}^{\infty}$ and $\{\mathcal{H}_j\}_{j=1}^{\infty}$ are series of continuous operator [30] applied on a finite dimensional space, the series of operators, $\{\mathcal{F}_{i,j}\}_{i,j=1}^{\infty}$, $\{\mathcal{H}_j\}_{j=1}^{\infty}$. And $\mathcal{F}_{i,j}: \mathbb{R}^{M \times N} \to \mathbb{R}^{M \times N}$. $\mathcal{H}_j: \mathbb{R}^{M \times N} \to \mathbb{R}^{M \times N}$. If we have: $\lim_{i \to \infty} \mathcal{F}_{i,j} \to \mathcal{H}_{M,j}$ and $\lim_{j \to \infty} \mathcal{H}_{M,j} \to \mathcal{G}$. Then, $\exists \lim_{k \to \infty} \mathcal{F}_{i_k, j_k} \to \mathcal{G}$ holds.

Theorem 4.1 demonstrates that a computational model satisfying Corollary 1.2 with multiple variables can converge to a fixed point via an alternative strategy. The proof of Theorem 1.2 can be viewed in Appendix A, Supplementary Material. Moreover, in Theorem 4.1, due to the convexity of the Augmented Lagrange function [32], each independent approximation, such as $\lim_{i \to \infty} \mathcal{F}_{i,j} \to \mathcal{H}_{M,j}$ and $\lim_{j \to \infty} \mathcal{H}_{M,j} \to \mathcal{G}$ can converge to a unique point.

# 3    Results

## 3.1    Comparison of Identified Hierarchical FCs via DEMAND and Other Peer Four Methods

We employ the resting-state fMRI signals from 15 healthy individuals in Consortium for Neuropsychiatric Phenomics (CNP) (https://openfmri.org/dataset/ds000030/) in order to validate the reconstruction performance of DEMAND. To avoid heterogeneous parameter tuning, all hyperparameters are tuned by following DEMAND's hyperparameter estimations. The estimated number of layers for DEMAND is two. The sizes of the first and second layers are 25 and 6, respectively. In addition, other parameters are tuned by following in [5-8]. The activation function of all layers of DEMAND and DBN is determined as Sigmoid. Furthermore, all abbreviations of templates can be found in Table S1, Appendix A, Supplementary Material.

The following figure shows the reconstruction of the first layer FCs via DEMAND and other peer methods, e.g., DICA, DNMF, SDDL, and DBN, compared with templates [33]. The qualitative validations demonstrate that the reconstruction of the first layer FCs of DEMAND and SDDL is comparable to original templates, especially when the intensity is considered.

Based on simple observations of Figures 1 and 2, the reconstruction of DEMAND is very similar to the original template/ground truth networks [33]. Furthermore, the intensity of the reconstructed FCs is very close to the templates. The theoretical explanation of the highest intensity matching of DEMAND is included in Appendix C. All representative slices of reconstructed FCs can be viewed via Figure S2, Appendix E, Supplementary Material.

Furthermore, the quantitative comparison of the first layer FCs is provided in Figure 3, where it compares the reconstructed first layer FCs of DEMAND with those of the other three peer methods based on original templates. The similarity is derived based on Hausdorff Distance [34]. In general, the intensity of FCs/features extracted via DBN and DEMAND is similar to the original templates.

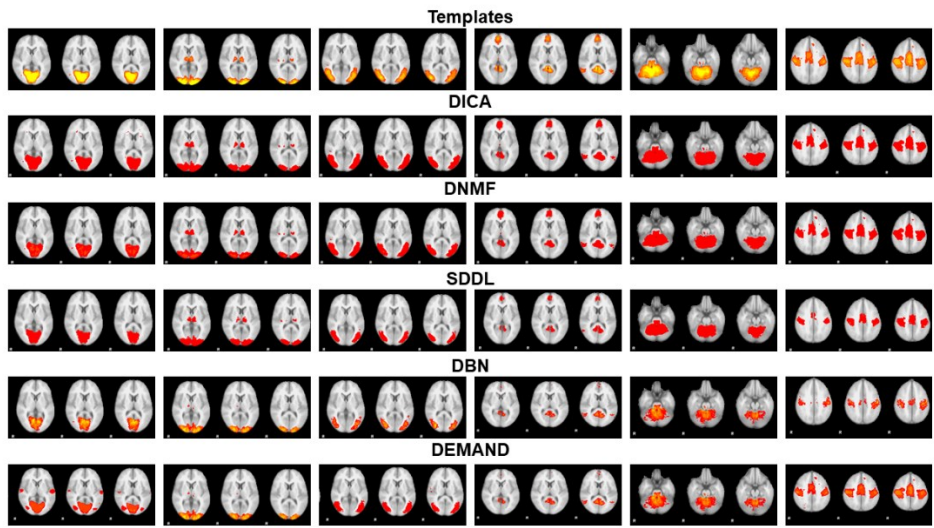

Figure 1. This figure presents six representative slices of reconstructed twelve FCs via DEMAND and other four peer methods.

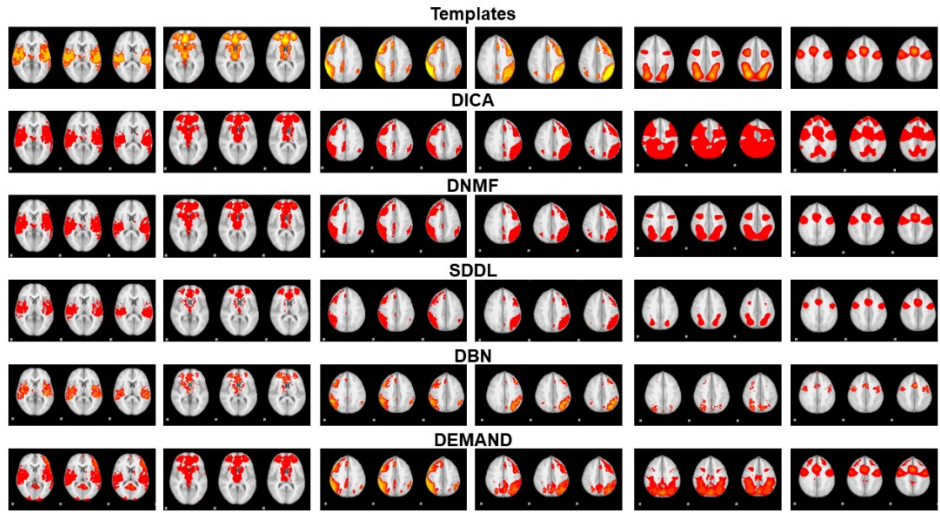

Figure 2. This figure presents another six representative slices of reconstructed twelve FCs via DEMAND and other three peer methods.

Furthermore, the following figure presents the reconstruction of the second FCs of DEMAND and the other four peer methods. Briefly, both linear and nonlinear models can successfully reconstruct hierarchical FCs. These higher-level FCs can be considered as the recombination of *shallow* FCs.

In detail, for instance, DEMAND can reconstruct FCs at the second layer with a strong spatial overlap with templates. Although other peer algorithms can reconstruct six FCs at the second layer, some revealed that FCs are significantly different from templates. Although other peer algorithms can reconstruct six FCs at the second layer, some revealed that FCs are significantly different from templates. Specifically, for example, in the FCs extracted by DICA in the first column, there has been an activated occipital lobe missed compared with templates; in addition, there is a significant difference between extracted FCs and templates, e.g., FCs in the first and second column; furthermore, the FCs identified by SDDL in third and fourth column demonstrates a significant disruption of areas to templates; similarly, DBN can only construct   Due to there has not been a 'ground-truth' to quantitatively validate the performance of DEMAND and other four pee methods, we investigate the reproducibility of hierarchical FC's identification instead.

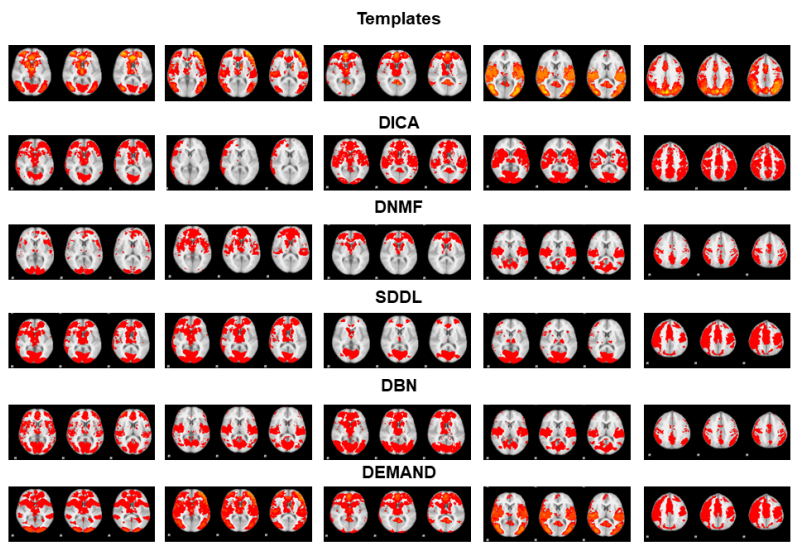

Figure 3. All qualitative comparison of reconstructed six FCs at second layer extracted via DEMAND and other four peer algorithms.

In Figure 3, DEMAND and DBN can reconstruct the second layer FCs more accurately compared to the templates; meanwhile, DICA and DNMF only reconstruct five and three FCs with higher similarity to the original templates, respectively. Furthermore, for SDDL, the reconstruction of FCs in the first and second columns is similar to original templates. Moreover, we also theoretically explain why DEMAND can provide more FCs/spatial features than Deep ICAs. Please view the proof in Theorem 4.2, Appendix D.

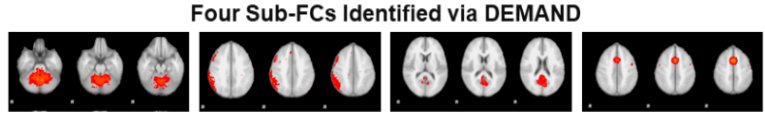

Figure 4. The presentation of sub-FCs, i.e., four FCs at the second layer, demonstrates minor/weaker patterns of FCs in deeper layers.

Nevertheless, using the same hyperparameters, other peer algorithms cannot reconstruct the minor FCs as DEMAND shown in Figures 3 and 4; furthermore, these individual/minor FCs would be more sensitive from a clinical translational perspective to a brain disease that can guide the personalized diagnosis and treatment. For instance, in the third column of Figure 4, a precuneus, a functional core of Default Mode Network (DMN) is detected. Furthermore, from a neuroscience perspective, these minor FCs could be more sensitive to some neurodegenerative diseases. Meanwhile, as discussed before, these sub-FCs also support the efficacy of the non-fully connected architectures of DEMAND which could avoid overfitting and enable the application of individual data sets to detect the potential minor FCs. These minor FCs are identified via DEMAND based on automatic hyperparameters tuning rather than the manually designed higher-order FCs provided by DICA [5].

Figure 5 (a) presents an example of an identified hierarchy of FCs via DEMAND. This hierarchy indicates the potential organizations of the meta, canonical and sub-FCs. In addition, to validate the reconstruction performance of DEMAND and the other three peer methods, the similarity between the identified FCs at the first layer and twelve ground-truth templates [32] is provided in Figure 5(b). In detail, all FCs extracted by DEMAND provide an overall higher similarity than four peer algorithms. Finally, to examine the reproducibility, we randomly separate the original input data into two independent sets as $FC_{test}$ and $FC_{retest}$ shown in Eqs. (9). The quantitative results of reproducibility are presented in Figure 5(c).

$$fc_i \in FC_{test}, fc_j \in FC_{retest}, FC_{test} \cap FC_{retest} = \emptyset \qquad (9)$$

$$reproduceVal \leftarrow corr(fc_i, fc_j)$$

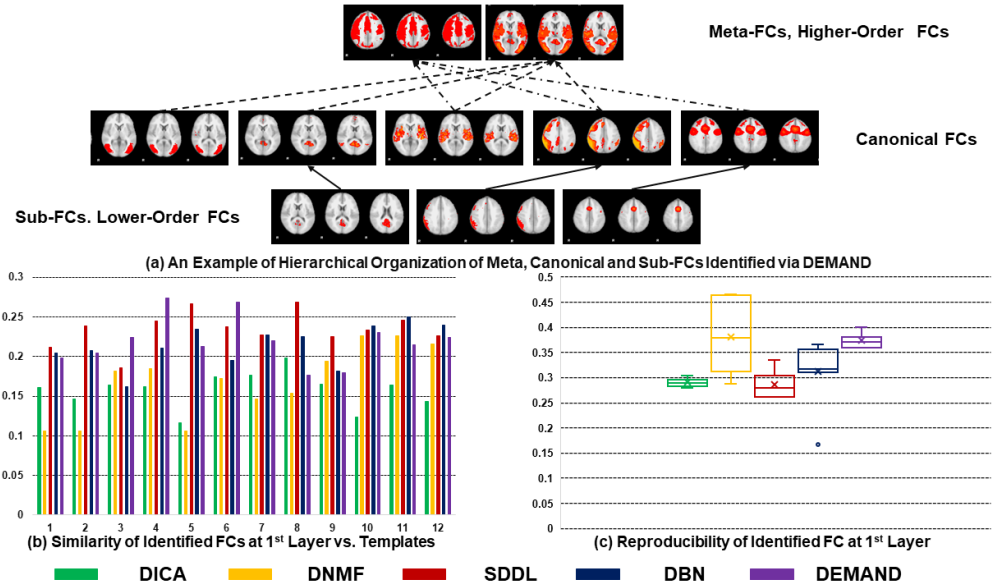

Figure 5. The presentation of higher-level FCs (2nd layer FCs) demonstrates the recombinations of the *shallow* FCs (1st layer FCs) in deeper layers.

Moreover, the following section employed the current released hierarchical templates to validate DEMAND and the other four peer methods.

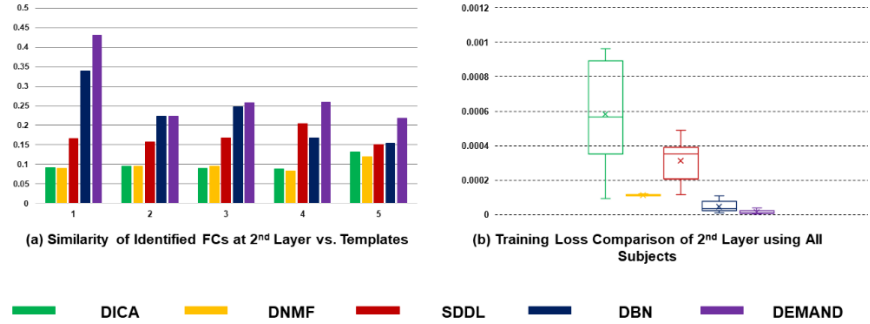

Figure 6. It demonstrates the reconstruction accuracy of all subjects' resting-state fMRI signals from healthy individuals in CNP (https://openfmri.org/dataset/ds000030/) via DEMAND and the other three peer methods.

In Figure 6 (a), similarly, the quantitative results demonstrate that the identified deeper features of DEMAND provide a higher reconstruction accuracy even compared with meta-FC templates reported in [40]. Moreover, the significant difference in reconstruction accuracy can be easily identified since the purple box plot in  shows the highest accuracy of training loss at the second layer.

## 4      Conclusion

In this work, DEMAND employs Adam and an alternative optimization strategy that is well suited to optimize convex/alternative convex problems and utilizes RRO for data-driven determination of all hyperparameters, which can be considered an advanced factorization method. Furthermore, the efficient non-fully connected architectures of DEMAND enable recognizing the individual/minor features, e.g., FCs, in individual brains without overfitting. In addition, the theoretical and experimental studies demonstrate that the reconstruction accuracy of DEMAND is comparable to or even better than DNNs.

In conclusion, DEMAND not only requires fewer training datasets but also can determine all hyperparameters automatically; according to validation results, DEMAND can also synergize research of neurodevelopmental, neurodegenerative, and psychiatric disorders since these revealed and novel biomarkers can benefit for diagnosis, prognosis and treatment monitoring [35-39],. Moreover, DEMAND provides an opportunity to explore the hierarchical spatial identifiability of FCs in order to generate the dominant hierarchy of FCs in the brain [40].

Overall, we believe that DEMAND can serve as an inspiring work for a fruitful future with the profound influence of deep nonlinear models.

# Supplemental Material

# Appendix A

***Definition 1.1*(Variance Bounded Real Function)** Given a real function $f$ denoted on $[a,b]$, and $\Delta: a = x_0 < x_1 < x_2 < \cdots < x_n = b$. A sum as $v_\Delta = \sum_{i=1}^{n}|f(x_i) - f(x_{i-1})|$ and $V_a^b(f) = \sup\{v_\Delta: \forall \Delta\}$. The variance bounded real function is denoted as $V_a^b(f) < \infty$.

***Definition 1.2* (Amplitude of Real Function)** Given a real function $f$ denoted on $[a,b]$, and $\forall B(x_0, \delta) \subseteq [a,b], \delta > 0$; $\omega_f(x_0) = \lim_{\delta \to 0} \sup\{|f(x') - f(x'')|: x', x'' \in B(x_0, \delta)\}$.

***Lemma 1.1* (Smooth & Variance Bounded Real Function)** If and only if a real function $f \in Lip1([a,b])$, $V_a^x(f) < \infty$ holds.

***Proof***: If $f \in Lip1([a,b])$, it indicates: $\forall x_1, x_2 \in [a,b]$, we have: $|f(x_1) - f(x_2)| \leq L|x_1 - x_2|$. Moreover, according to Definition 1, we have:

$$v_\Delta = \sum_{i=1}^{n}|f(x_i) - f(x_{i-1})| \leq L(|x_0 - x_1| + |x_1 - x_2| + \cdots + |x_n - x_{n-1}|) \quad (A.1)$$
$$\leq L(b-a) < \infty$$

If $V_a^x(f) < \infty$ holds, it demonstrates: $\sup\{v_\Delta: \forall \Delta\} < \infty$, let $x_n$ be $x$, we have:

$$\sum_{i=1}^{n}|f(x_i) - f(x_{i-1})| \leq \sup\{v_\Delta: \forall \Delta\} < \infty \quad (A.2)$$

Furthermore, if $n \to \infty$, and $\sum_{i=1}^{n}|f(x_i) - f(x_{i-1})| < \infty$ holds, obviously, we must have:
$$|f(x_i) - f(x_{i-1})| \to 0 \quad (A.3)$$
It should satisfy:
$$x_i, x_{i-1} \in B(x, \varepsilon), \forall \varepsilon > 0 \; |f(x_i) - f(x_{i-1})| \leq L|x_i - x_{i-1}| \quad (A.4)$$
Since $\forall x_i, x_{i-1} \in [a,b]$, obviously, we have:
$$f \in Lip1([a,b])$$

***Lemma 1.2* (Amplitude & Variance Bounded Real Function)** $\omega_f(x_0) = \lim_{\delta \to 0} \sup\{|f(x') - f(x'')|: x', x'' \in B(x_0, \delta) \subseteq [a,b]\} < \varepsilon$ is equivalent to $f \in Lip1([a,b])$.

***Proof***: If $f \in Lip1([a,b])$ holds, similarly, if $n \to \infty$, $\forall \{x_i\}_{i=1}^n$, we have
$$|f(x_i) - f(x_{i-1})| \to 0 \quad (A.5)$$
Replace $x_i$ and $x_{i-1}$ by $x'$, $x''$, respectively, it satisfies:
$$\omega_f(x_0) = \lim_{\delta \to 0} \sup\{|f(x') - f(x'')|: x', x'' \in B(x_0, \delta) \subseteq [a,b]\} < M \quad (A.6)$$
If $\omega_f(x_0) = \lim_{\delta \to 0} \sup\{|f(x') - f(x'')|: x', x'' \in B(x_0, \delta) \subseteq [a,b]\} < \infty$, we assume:
$$\omega_f(x_0) = \lim_{\delta \to 0} \sup\{|f(x') - f(x'')|: x', x'' \in B(x_0, \delta) \subseteq [a,b]\} < \varepsilon \quad (A.7)$$
Obviously, given $x', x'' \in B(x_0, \delta) \subseteq [a,b]$, we have:
$$\forall \varepsilon > 0 \; |f(x') - f(x'')| < \varepsilon \quad (A.8)$$
When $\delta \to 0$, let $|x' - x''| = \frac{\varepsilon}{L}$, it also indicates:

$$|f(x') - f(x'')| < L|x' - x''| \qquad (A.9)$$

***Lemma 1.3* (Cantor Theorem)** Given $\{B_i\}_{i=1}^{\infty}$ are closed sets and $\forall B_i \neq \emptyset$, if $B_1 \supseteq B_2 \supseteq \cdots \supseteq B_k \supseteq \cdots$, $\cap_{i=1}^{\infty} B_i \neq \emptyset$.

***Lemma 1.4* (Vitali Covering Lemma)** Given $\{B_i\}_{i=1}^{n}$ are closed sets and $\forall B_i \cap B_j = \emptyset$, $i \neq j$, $E \subseteq \mathbb{R}$, and $m^*(E) < \infty$, if $m^*(E \setminus \cup_{i=1}^{n} B_i) < \varepsilon$, $\forall \varepsilon > 0$, holds, $\{B_i\}_{i=1}^{n}$ defines a Vitali Covering of $E$.

**Lemma 1.5 (Heine-Borel Covering Theorem)** Given $\Gamma$ is a close and bounded set. Then, an open set sequence as $\{g_i\}_{i=1}^{K} \stackrel{\text{def}}{=} G$, $\cup_{i=1}^{K} g_i \supseteq \Gamma$, and $\bar{\bar{G}} = \aleph_0$.

**Lemma 1.6 (Composition of Function)** Given $g \in Lip1([a,b])$, and $g$ is not a constant real function, if $f \notin Lip1([a,b])$, $g(f(x)) \notin Lip1([a,b])$ holds.
***Proof***: Proof by contradiction, if assume $g(f(x)) \in Lip1([a,b])$, $\forall x_1, x_2 \in [a,b]$, $f(x_1), f(x_2) \in [a,b]$,
$$|g(f(x_1)) - g(f(x_2))| < L_g|f(x_1) - f(x_2)| < N < \infty \text{ and } L_g \neq 0. \qquad (A.10)$$
However, since $f \notin Lip1([a,b])$, we have: $|f(x_1) - f(x_2)| > M$ that is contradiction with $|f(x_1) - f(x_2)| < \frac{N}{L_g}$. Thus, $(f(x)) \notin Lip1([a,b])$ holds.

***Theorem 1.1* (Composition of Non-Smooth Activation Functions)** Given a non-smoothed activation function $f_i$ with a single non-smooth point, denoted on $[a,b] \subseteq \mathbb{R}^1$, $f_i \in Lip1([a,b]\setminus\{x_i\})$ $i \in \mathbb{N}$. And the composition of $f_i$ and $f_j$, represented as $f_{j,i} \stackrel{\text{def}}{=} f_j(f_i(x))$, results in $\mathcal{F} \stackrel{\text{def}}{=} f_{\cdots,k,\cdots j,i} \in Lip1([a,b]\setminus[c,d])$, when $k \to \infty$, and $m([c,d]) \neq 0$; moreover, given $t \to \infty$, the summation as $\sum_{i=1}^{t} \mathcal{F}_t$ leads to $\sum_{i=1}^{t} \mathcal{F}_t \notin Lip1([a,b]\setminus[c',d'])$ and $m([c',d']) \neq 0$. $m(\cdot)$ represents the Lebesgue measure.
***Proof***: At first, we discuss $k < \infty$, and we assume, $f \in Lip1([a,b]\setminus\{x_0\})$
According to Lemma 1 and Lemma 2, if $x_0 \in B(x_0, \delta)$, for we have:
$$\omega_{f_i}(x_i) = \lim_{\delta \to 0} \sup\{|f_i(x') - f_i(x'')| : x', x'' \in B(x_i, \delta) \subseteq [a,b]\} > M \qquad (A.10)$$

$$\omega_{f_j}(x_j) = \lim_{\delta \to 0} \sup\{|f_j(y') - f_j(y'')| : y', y'' \in B(x_j, \delta) \subseteq [a,b]\} > M \qquad (A.11)$$
Thus, we have $f_i$ and $f_j$ are not smooth on $B(x_i, \delta)$ and $B(x_j, \delta)$, respectively.
And for the composition, let $k = 2$,
$$\omega_{f_{j,i}}(x_i) = \lim_{\delta \to 0} \sup\{|f_j(f_i(x')) - f_j(f_i(x''))| : x', x'' \in B(x_i, \delta) \subseteq [a,b]\} \qquad (A.12)$$
Let $f_i(x') = x_j'$ and $f_i(x'') = x_j''$, if $(x_j', x_j'') \in B(x_k, \delta)$, it is easy to prove the amplitude of $f_{j,i}$ as following:
$$\omega_{f_{j,i}}(x_1) = \lim_{\delta \to 0} \sup\{|f_2(x_j') - f_2(x_j'')| : x_j', x_j'' \in B(x_k, \delta) \subseteq B(x_j, \delta) \subseteq [a,b]\} > M \qquad (A.13)$$
Naturally, we need to analyze other relations of $B(x_k, \delta)$ and $B(x_j, \delta)$; in detail, there are five situations to be discussed separately:
1). Assume, if $\forall B(x_k, \delta) \cap B(x_j, \delta) = \emptyset$, $i, j \in \mathbb{R}$, $i \neq j$, obviously, due to the same composition, we have: $B(x_k, \delta) \cap B(x_{k-1}, \delta) = \emptyset$, according to Lemma 1.3,
$$[c,d] \setminus \bigcup_{k=1}^{\infty} B(x_k, \delta) = \{\hat{x}_j\}_{j=1}^{N} \qquad (A.14)$$
And

$$m\left(\{\hat{x}_j\}_{j=1}^N\right) = 0 \tag{A.15}$$

According to Lemma 1.6, $f_j(f_i(B(x_i,\delta)) \notin Lip1(B(x_k,\delta) \cup B(x_j,\delta))$, therefore, we have:
$$f_{\cdots,k,\cdots j.i} \notin Lip1([c,d]\setminus\{\hat{x}_j\}_{j=1}^N) \tag{A.16}$$

2). Similarly, if we assume $\forall B(x_k,\delta) \cap B(x_j,\delta) \neq \emptyset$, due to the composition, we have: $B(x_k,\delta) \cap B(x_{k-1},\delta) \neq \emptyset$, according to Lemma 1.5,
$$[a,b] \supseteq \bigcup_{k=1}^K B(x_k,\delta) \supseteq [c,d] \tag{A.17}$$

Therefore, we can conclude:
$$f_{\cdots,k,\cdots j.i} \notin Lip1([c,d]) \tag{A.18}$$

3). Moreover, if $B(x_j,\delta) \supseteq B(x_k,\delta) \supseteq B(x_{k+1},\delta) \supseteq \cdots$, according to Lemma 1.5, we have:
$$\cap_{i=1}^K B(x_i,\delta) = \Xi \neq \emptyset \tag{A.19}$$

It means:
$$f_{\cdots,k,\cdots j.i} \in Lip1([a,b]\setminus B(x_j,\delta)) \tag{A.20}$$

Thus, similarly, we have:
$$f_{\cdots,k,\cdots j.i} \in Lip1([a,b]\setminus\Xi) \tag{A.21}$$

4). Finally, if $\cdots \supseteq B(x_{k+1},\delta) \supseteq B(x_k,\delta) \supseteq B(x_j,\delta)$,
Therefore, based on (1) and (2), we have:
$$\omega_{f_{k,\cdots,2,1}}(x_2) = \lim_{\delta \to 0} \sup\{|f_{k,\cdots,2,1}(x') - f_{k,\cdots,2,1}(x'')|: x',x'' \in [c,d]\} > M \tag{A.21}$$

It indicates:
$$f_{\cdots,k,\cdots j.i} \in Lip1([a,b]\setminus\bigcup_{i=1}^\infty B(x_i,\delta)) \tag{A.22}$$

5). Comprehensively, the situation includes all previously discussed (1) to (4), it is easy to conclude:
$$f_{\cdots,k,\cdots j.i} \in Lip1([a,b]\setminus[c,d] \tag{A.23}$$

Using Lemma 1.1, obviously, given $\Delta: a = x_0 < x_1 < x_2 < \cdots < x_n = b$, and $\Delta': x' < \hat{x}_1 < \hat{x}_2 < \cdots < \hat{x}_n < x''$, $v_{\Delta_1} + v_{\Delta_2} = v_\Delta$.

$$v_\Delta + v_{\Delta'} = v_{\Delta_1} + v_{\Delta_2} + v_{\Delta'}$$
$$= \sum_{i=1}^{n_1} |f(x_i) - f(x_{i-1})| + \sum_{i=n_1}^{n_2} |f(x_i) - f(x_{i-1})| \tag{A.24}$$
$$+ \sum_{i=n_2}^{n} |f(x_i) - f(x_{i-1})|$$

Since $v_{\Delta'} > M$, $v_\Delta + v_{\Delta'} > M$, it is easy to have:
$$\sum_{i=1}^t f_i^t \notin Lip1([a,b]) \tag{A.25}$$

Table S1. All abbreviations of canonical FCs in Methodological Validation

| Name/ | Number/Abbreviation | Name | Abbreviation |
|---|---|---|---|
| Primary Visual Network | 1/VIS-1 | Auditory Network | 7/AUD |
| Perception Visual Shape Network | 2/VIS-2 | Executive Control Network | 8/ECN |
| Perception Visual Motion Network | 3/VIS-3 | Left Frontoparietal Network | 9/FP-L |
| Default Mode Network | 4/DMN | Right Frontoparietal Network | 10/FP-R |
| Brainstem & Cerebellum Network | 5/BC | Dorsal Attention Network | 11DAN |
| Sensorimotor Network | 6/SM | Salience Network | 12/SN |

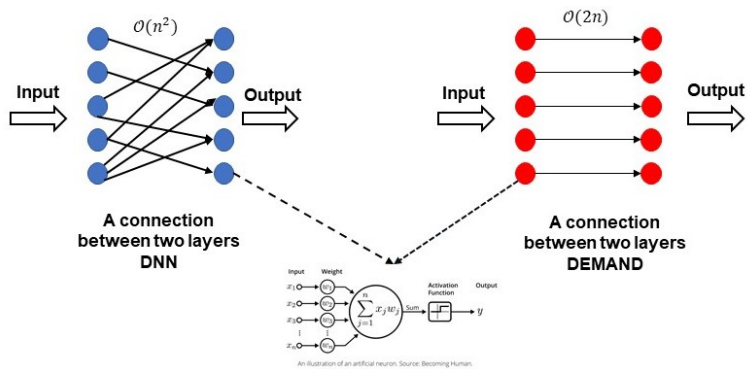

**Figure S2**. An example demonstrates fully connected structures used by DNN and non-fully architecture used by DEMAND. Obviously, the DMAND's architecture is more efficient, which is easy to be optimized and not vulnerable to overfitting.

# Appendix B

**Algorithm 2.1 (Core Algorithm):** Deep Matrix Approximately Nonlinear Decomposition (DEMAND)

**Input:** $I \in \mathbb{R}^{t \times m}$, $SG$ is the input signal matrix; set $\lambda > 1$ as the penalty parameter; randomly initialize $\{X_k\}_{k=1}^{M}$, $\{Y_k\}_{k=1}^{M}$ and $\{S_k\}_{k=1}^{M}$ ;
Set $r$ as the initial estimated rank of $X_1$ and $Y_1$ and layer $k$ as 0.
   *while* rank > 1
      update $X_k$ using Eq. (3-1);
      update $Y_k$ using Eq. (3-2);
      update $S_k$ using Eq. (3-3);
      apply sparse operator on $S_k$;
      update multiplier $e_k$ using Eq. (3-4);
      use **Algorithm 2.2** to estimate rank of $Y_k$;
      $k \leftarrow k + 1$;
  *end while*
$K \leftarrow k$;
*Use Algorithm 2.5 to perform matrix back propagation;*
**Output:** $\{X_i\}_{i=1}^{K} \in \mathbb{R}^{n \times m}$, $\{Y_i\}_{i=1}^{K} \in \mathbb{R}^{n \times m}$ and $\{S_i\}_{i=1}^{K} \in \mathbb{R}^{n \times m}$ ;

---

**Algorithm 2.2:** Rank Reduction Operator (RRO)

**Input:** $Y_k \in \mathbb{R}^{n \times m}$, $Y_k$ is the feature matrix;
  $YR_k \leftarrow QR(Y_k)$;
  $minRank \leftarrow \min(1, size(Y_k))$;
  $estRank \leftarrow minRank - 1$;
  $diagYR_k \leftarrow abs(diag(YR_k))$;
  using **Algorithm 2.3** calculate the weighted difference of $diagYR_k$;
  set weighted difference of $diagYR_k$ as $wd$;
  $[rankMax1, posMax1] \leftarrow \max(wd)$;
  using **Algorithm 2.4** to calculate the weighted ratio of $diagYR_k$;
  set weighted ratio of $diagYR_k$ as $wr$;
  $[rankMax2, posMax2] \leftarrow \max(wr)$;
  *if* rankMax1 is equal to 1
    $estRank \leftarrow posMax1$;
  *end if*
  $valWR \leftarrow \text{find}(wr > rankMax2)$;
  *if* number (valWR) is equal to 1
    $estRank \leftarrow posMax1$;
  *end if*
  using **Algorithm 2.5** to calculate the weighted correlation of $diagYR_k$;
  set weighted correlation of $diagYR_k$ as wc;
  $[rankMax3, posMax3] \leftarrow \max(wc)$;

 ***if*** rankMax1 is equal to 1
   $estRank \leftarrow posMax3$;
 ***end if***
 *valWC*←find( *wc>rankMax3* );
 ***if*** number (valWC) is equal to 1
   $estRank \leftarrow posMax3$;
 ***end if***
 $estRank \leftarrow \max(posMax1, posMax2, posMax3)$;
***Output: estRank***

---

**Algorithm 2.3:** Weighted Difference (WD)

***Input:*** $Vec \in \mathbb{R}^{1 \times m}$, *Vec* is a vector;
 *cumSum* ← Calculate *cumulative sum of Vec;*
 *diffVec* ← Calculate differences between adjacent elements of *Vec*;
 *resverseVec* ← reverse *Vec*;
 *WD* ← abs(*diffVec*) ./ *reverseVec*;
***Output:*** *WD*

---

**Algorithm 2.4:** Weighted Ratio (WR)

***Input:*** $Vec \in \mathbb{R}^{1 \times m}$, *Vec* is a vector;
 $L$ ← Calculate length of *Vec*;
 *ratioVec* ← *Vec(1:L-1)* ./ *Vec(2:L)*;
 *WR* ← (L-2)\**ratioVec* ./ sum (*ratioVec*);
***Output:*** *WR*

---

**Algorithm 2.5:** Weighted Correlation (WC)

***Input:*** $Vec \in \mathbb{R}^{1 \times m}$, *Vec* is a vector;
*WC*←Calculate the weight correlation using Eq. (7)
***Output:*** *WC*

---

**Algorithm 2.6:** Matrix Backpropagation (MBP)

***Input:*** $\{X_k\}_{k=1}^M$, $\{Y_k\}_{k=1}^M$, $I$, and set $E = 0.01$, MaxIter
$$\{X_k^1\}_{k=1}^M \leftarrow \{X_k\}_{k=1}^M$$
$$\{Y_k^1\}_{k=1}^M \leftarrow \{Y_k\}_{k=1}^M$$
***for*** *it* in 1 to *MaxIter*
$$K \leftarrow (\prod_{k=1}^{M-1} X_k^{it})^T \cdot I$$
$$P_k^{it} \leftarrow (X_k^{it})^T X_k^{it} \cdot \prod_{k=1}^{M-1} \max(X_k)$$
$$c_k^{it} \leftarrow \max(X_k) \cdot \frac{d\mathcal{N}^{-1}(x)}{dx}, x = X_k^{it} Y_k^{it}$$
$$V_k^{it} \leftarrow (X_k^{it})^T \cdot (P_k^{it} \cdot \mathcal{N}^{-1}(x) - K) \odot c_k^{it}, x = X_k^{it} Y_k^{it}$$
$$Y_k^{it+1} \leftarrow Y_k^{it} - \frac{E}{2^{it}}(V_k^{it})$$

$$X_k^{it+1} \leftarrow X_k^{it+1} - \frac{E}{2^{it}}(V_k^{it})$$

*end*
***Output:*** $\{X_k\}_{k=1}^M$ and $\{Y_k\}_{k=1}^M$

# Appendix C

The matrix polynomials are defined as: $\forall k \in \mathbb{N}\ P_{2k}(X) = (XX^T)^k$, $P_{2k+1}(X) = (XX^T)^k X$, $X \in \mathbb{R}^{S \times T}$; and $\{P_n(X)\}_{n=1}^N$ defines a series of matrix polynomials, for example: $\{P_n(X)\}_{n=1}^3 = \{X, XX^T, XX^T X\}, X \in \mathbb{R}^{S \times T}$;

Moreover, it is easy to prove $\{P_n\}_{n=1}^\infty$ denoted on $\mathbb{R}^{S \times T}$ as a ring $(\{P_n\}_{n=1}^\infty, +, \times)$, and it also demonstrates $\{P_n(X)\}_{n=1}^N \in (\{P_n\}_{n=1}^\infty, +, \times)$, e.g., $(\{P_n\}_{n=1}^\infty, +, \times) \supseteq \sum_{i=1}^M \{P_n(X)\}_{n=1}^N$. (Dummit, 2004; Kadison, 1997). Then we introduce the theorem 1.1 to describe the superiority of DEMAND.

***Theorem 3.1*** **(Superiority of DEMAND)** Given a real function $f: \mathbb{R}^{S \times T} \to \mathbb{R}^{S \times T} \cup \{\pm \infty\}$ and $m(\{Y \subseteq \mathbb{R}^{S \times T}: |f(Y)| = \pm\infty\}) = 0$. And $m(\cdot)$ represents the Lebesgue measure. If considering $X \in \mathbb{R}^{S \times T}$ and the series of matrix polynomials $\{\mathcal{N}_n(X)\}_{n=1}^N$, we have: if $N$ is large enough, we have: $\forall\ \varepsilon > 0\ \|\{\mathcal{N}_n(X)\}_{n=1}^N - f(X)\| \leq \varepsilon$.

***Proof***: According to Лузин (Luzin) Theorem (Royden, 1968), we have a close set:
$$F_n \subset F_{n+1} \subset \cdots \subseteq \mathbb{R}^{S \times T}$$
$$m(\mathbb{R}^{S \times T} \setminus F_k) = \frac{1}{k}, k \in \mathbb{N} \tag{C.1}$$
$$f \in C(F_k)$$

Then we have a consistent real function $g(X)$, and obviously we have:
$$g(X) = f(X) \tag{C.2}$$

Since for any continuous real function, we have:
$$|g(X) - P_k(X)| < \frac{1}{k} \tag{C.3}$$

Let $\mathcal{F} = \bigcup_{k=1}^\infty F_k$, and obviously we have:
$$m(\mathbb{R}^{S \times T} \setminus F_k) = m(\mathbb{R}^{S \times T} \setminus \bigcup_{k=1}^\infty F_k) = \bigcap_{k=1}^\infty m(\mathbb{R}^{S \times T} \setminus F_k) = \bigcap_{k=1}^\infty \frac{1}{k} = 0 \tag{C.4}$$

Moreover, it is easy to prove $\{P_n\}_{n=1}^\infty$ denoted on $\mathbb{R}^{S \times T}$ as a ring $(\{P_n\}_{n=1}^\infty, +, \times)$, and it also demonstrates $\{P_n(X)\}_{n=1}^N \subseteq (\{P_n\}_{n=1}^\infty, +, \times)$, e.g., $P_n(X) \stackrel{\text{def}}{=} \prod_{i=1}^N x_i + \sum_{j=1}^N y_j$ (Dummit, 2004; Kadison, 1997).

If $\mathfrak{F}$ is a real function denoted on set $\mathcal{F}$, it indicates:
$$\lim_{N \to \infty} |\mathfrak{F} - \{P_n(X)\}_{n=1}^N| = 0 \tag{C.5}$$
then we have $\lim_{N \to \infty} \{P_n(X)\}_{n=1}^N = \mathfrak{F}$, meanwhile, if $N$ is large enough, $|\mathfrak{F} - \{P_n(X)\}_{n=1}^N| < \varepsilon$ holds.

Moreover, if $\{\mathcal{N}_n(X)\}_{n=1}^N \in C(\mathbb{R}^{S \times T})$, according to Theorem, $|\mathcal{N}_n(X) - \{P_n(X)\}_{n=1}^\infty| \to 0$; we have:
$$|\mathfrak{F} - \{\mathcal{N}_n(X)\}_{n=1}^N| = |\mathfrak{F} - \{P_n(X)\}_{n=1}^N| < \varepsilon \tag{C.6}$$

***Theorem 3.2*** **(Random Initialization Operator is bounded)** If we denote the sparse operator as $\mathcal{I}: \mathbb{R}^{S \times T} \to \mathbb{R}^{S \times T}$, we have $\|\mathcal{M}\| < \infty$.

***Proof***: according to the definition of operator norm (Rudin 1973), $\|\mathcal{I}\| \leq \sup \frac{\|MX\|}{\|X\|}$; obviously, $\|\mathcal{I}X\|$ and $\|X\|$ is bounded, since both of norms are based on finite dimensional matrix. And if we denote:

$$X = \begin{bmatrix} a_1 \\ a_2 \\ \vdots \\ a_{n-1} \\ a_n \end{bmatrix}\ \mathcal{I}X = \begin{bmatrix} b_1 \\ b_2 \\ \vdots \\ b_{n-1} \\ b_n \end{bmatrix}\ \|X\| < \infty\ \|\mathcal{I}X\| < \infty \tag{C.7}$$

Obviously, $\|\mathcal{I}\| < \infty$.

***Theorem 3.3*** **(Sparsity Operator is Contraction)** If we denote the sparse operator as $\mathcal{S}: \mathbb{R}^{S \times T} \to \mathbb{R}^{S \times T}$, we have $\|\mathcal{S}\| < \infty$.

***Proof***: according to the definition of operator norm (Rudin, 1973), $\|\mathcal{S}\| \leq sup\frac{\|\mathcal{S}X\|}{\|X\|}$; obviously, $\|\mathcal{S}X\|$ and $\|X\|$ is bounded, since both of norms are based on finite dimensional matrix. And if we denote:

$$X = \begin{bmatrix} a_1 \\ a_2 \\ \vdots \\ a_{n-1} \\ a_n \end{bmatrix} \mathcal{S}X = \begin{bmatrix} a_1 \\ 0 \\ \vdots \\ a_{n-1} \\ a_n \end{bmatrix} \tag{C.8}$$

Without loss of generality, and based on Lemma 1.2, we calculate the $\ell_2$ norm, and we have:

$$s = \|\mathcal{S}\| \leq sup\frac{\|\mathcal{S}X\|}{\|X\|} = \frac{\sum_{i=1}^{k}(a_i)^2}{\sum_{i=1}^{n}(a_i)^2} \tag{C.9}$$

Since $k < n$,

$$s = \|\mathcal{S}\| < 1 \tag{C.10}$$

This inequality demonstrates that $\|\mathcal{S}\|$ is contraction operator.

***Theorem 3.4 (Rank Reduction Operator is bounded)*** If we denote the sparse operator as $\mathcal{R}: \mathbb{R}^{S \times T} \to \mathbb{R}^{S \times T}$, we have $\|\mathcal{R}\| < \infty$.

***Proof***: According to the definition of operator norm (Rudin, 1973), $\|\mathcal{R}\| \leq sup\frac{\|\mathcal{R}X\|}{\|X\|}$; obviously, $\|\mathcal{R}X\|$ and $\|X\|$ is bounded, since both of norms are based on finite dimensional matrix. And if we denote:

$$X = \begin{bmatrix} a_1 \\ a_2 \\ \vdots \\ a_{n-1} \\ a_n \end{bmatrix}, \mathcal{R}X = \begin{bmatrix} a_1 \\ a_2 \\ \vdots \\ a_k \\ \vdots \\ a_{n-1} \end{bmatrix} \tag{C.11}$$

Eq. (C11) implies:

$$sup\frac{\|\mathcal{R}X\|}{\|X\|} = \frac{\sum_{i=1}^{n} a_i^2}{\sum_{i=u}^{p}(a_i - b_i)^2 + \sum_{i=v}^{q} a_i^2} < \infty \tag{C.12}$$

Also, if we examine the weighted ratio and weight difference, only considering the finite dimensional space, we have:

$$X = \begin{bmatrix} a_1 \\ a_2 \\ \vdots \\ a_{n-1} \\ a_n \end{bmatrix}, WR \cdot X = \begin{bmatrix} a_2/a_1 \\ a_3/a_2 \\ \vdots \\ a_k/a_{k-1} \\ \vdots \\ a_n/a_{n-1} \; 0 \end{bmatrix} WD \cdot X = \begin{bmatrix} a_2 - a_1 \\ a_3 - a_2 \\ \vdots \\ a_k - a_{k-1} \\ \vdots \\ a_n - a_{n-1} \end{bmatrix}$$

Obviously, for each rank estimation, the dimension of input matrix can be reduced at least by one. Similarly, WR and WD can be considered as the contract operators for dimensional estimation. It demonstrates that the input matrix can be reduced to a vector by *n-1* iterations at most.

***Lemma 3.1 (Contraction of Operators Combination)*** Given two contraction mappings $\Phi_1$ and $\Phi_2$, we have the composite of two contraction mapping as $\Phi_2 \cdot \Phi_1$. The composite mapping $\Phi_2 \cdot \Phi_1$ must be contractive.

***Proof***: According to the definition of contraction linear operator, we have:

$$\exists \zeta \in (0,1)$$
$$\rho \stackrel{\text{def}}{=} \|\Phi x - \Phi y\| \tag{C.13}$$
$$\rho(\Phi x, \Phi y) \leq \zeta \rho(x, y)$$

Obviously, and we have:

$$\rho(\Phi_1 u, \Phi_1 v) \leq \zeta \rho(u, v) \; \forall \zeta \in (0,1) \tag{C.14}$$
$$\rho(\Phi_2 x, \Phi_2 y) \leq \eta \rho(x, y) \; \forall \eta \in (0,1)$$

If we set:

$$x = \Phi_1 u, y = \Phi_1 v \tag{C.15}$$

the inequality below holds:
$$\rho(\Phi_2 x, \Phi_2 y) \leq \eta \rho(\Phi_1 u, \Phi_1 v) \leq \zeta \eta \rho(u, v) \tag{C.16}$$

Since the definition as
$$\forall \zeta, \eta \in (0,1), \rho(\Phi_2 \Phi_1 u, \Phi_2 \Phi_1 y) \leq \zeta \eta \rho(u, v) \tag{C.17}$$

***Lemma 3.2* (Adam Operator is bounded)** [26] If we denote the Adam optimizer operator as $\mathcal{A}: \mathbb{R}^{S \times T} \to \mathbb{R}^{S \times T}$, we have $\|\mathcal{A}\| < \frac{1}{\sqrt{T}}$.

***Corollary 3.1* (General Contraction Operator)** According to Lemma 1.2, if denote the operators $\{\Phi_i\}_{i=1}^K$, $\forall \Phi_i\ i \in \mathbb{N}$, $\Phi_i: \mathbb{R}^{S \times T} \to \mathbb{R}^{S \times T}$; considering any combination of operators: $\Phi_K \cdot \cdots \cdot \Phi_2 \cdot \Phi_1$, if at least a single operator $\Phi_i$ is contraction operator, and other operators are bounded, such as $\forall i \neq k\ \|\Phi_i\| \leq M$. If and only if $\prod_{i=1}^K \|\Phi_i\| < 1$, the combination of operator series $\Phi_K \cdot \cdots \cdot \Phi_2 \cdot \Phi_1$ is a contraction operator.

***Proof***: Obviously, according to Lemma 1.2, use a series as $\{\zeta_i\}_{i=1}^K$ to replace $\zeta, \eta \in (0,1)$,
Obviously, we have:
$$\zeta_i \in (0,1)\ i \in \mathbb{N} \tag{C.18}$$
$$\rho(\Phi_K \cdot \cdots \cdot \Phi_2 \Phi_1 u, \Phi_K \cdot \cdots \cdot \Phi_2 \Phi_1 y) \leq \zeta_K \cdot \cdots \zeta_2 \cdot \zeta_1 \cdot \rho(u,v)$$

Since $\zeta_K \cdot \cdots \zeta_2 \cdot \zeta_1 < 1$, we have proved this corollary.

***Corollary 3.2* (Iterative Contraction Operator)** According to Lemma 1.2, if denote the operators $\{\Phi_i\}_{i=1}^K$, $\forall \Phi_i\ i \in \mathbb{N}$, $\Phi_i: \mathbb{R}^{S \times T} \to \mathbb{R}^{S \times T}$; considering any combination of operators: $\Phi_K \cdot \cdots \cdot \Phi_2 \cdot \Phi_1$, if at least a single operator $\Phi_i$ is contraction operator, and other operators are bounded, such as $\forall i \neq k, \|\Phi_i\| \leq M$. If and only if $\lim_{n \to \infty} \prod_{i=1}^K \|\Phi_i\|^n = c < 1$, the combination of operator series $\Phi_K^n \cdot \cdots \cdot \Phi_2^n \cdot \Phi_1^n$.

***Proof***: Obviously, according to Lemma 3.1 and Corollary 3.1, use a series as $\{\zeta_i\}_{i=1}^K$ to replace $\zeta, \eta \in (0,1)$,
And we have:
$$\forall \zeta_i \in (0,1)\ i \in \mathbb{N} \tag{C.19}$$
$$\rho(\Phi_K^n \cdot \cdots \cdot \Phi_2^n \cdot \Phi_1^n u, \Phi_K^n \cdot \cdots \cdot \Phi_2^n \cdot \Phi_1^n y) < \zeta_i^n \cdot \cdots \cdot \zeta_2^n \cdot \zeta_1^n \cdot \rho(u,v)$$
Since $0 < \zeta_i^n \cdot \cdots \cdot \zeta_2^n \cdot \zeta_1^n < 1$, we have proved this corollary.

***Theorem 3.5* (Convergence of DEMAND in Finite Dimensionality Space)** DEMAND can converge as fast as Adam.

***Proof***: If we have $U, V \in \mathbb{R}^{m \times n}$, according to Theorems 3.2-3.4, and Lemma 3.2, the DEMAND can be represented as
$$DEMAND \stackrel{\text{def}}{=} (\mathcal{SRNA})^k \cdot \mathcal{I}: \mathbb{R}^{t \times m} \to \mathbb{R}^{t \times m} \tag{C.19}$$
According to Lemma 3.2, Corollary 3.1 and 3.2, we conclude:
$$\|\mathcal{A}^k U - \mathcal{A}^k V\| \leq \rho^k \|U - V\| \tag{C.20}$$
and $0 < \rho^k < 1$ holds and $\rho^k$ could be equal to $\mathcal{O}(\frac{1}{\sqrt{T}})$.

And given other definitions of operators adopted by DEMAND, obviously, since all norms of these operators are bounded, the following norm inequality holds:
$$\|(\mathcal{SRNA})^k \mathcal{I} \cdot U - (\mathcal{SRNA})^k \mathcal{I} \cdot V\| \leq (\frac{N}{asr})^k \cdot \|U - V\| \tag{C.20}$$

If $0 < (\frac{N}{asr})^k < 1$, $k \to \infty$, can guarantee the convergence of DEMAND comparable to DEMAND. It demonstrates that the convergence velocity of DEMAND would be equal to Adam, since the convergence velocity of Adam has been proved as $\mathcal{O}(\frac{1}{\sqrt{T}})$, and $(\frac{N}{asr})^k$ can be rewritten as $\frac{1}{\sqrt{asr}} (\frac{N}{asr})^k \sqrt{asr}$, if and only if $(\frac{N}{asr})^k \sqrt{asr}$ is a constant $\mathcal{C}$, and let $\frac{1}{\sqrt{asr}}$ be $\frac{1}{\sqrt{T}}$.

*Collaroy 3.3* **(Convergence of DEMAND in Infinite Dimensionality Space)** Given the infinite dimensionality space, the DEMAND is denoted as an operator as $\mathcal{D}: \mathbb{R}^{\infty \times \infty} \to \mathbb{R}^{m \times n}$. If we assume $\mathcal{D}$ and $\mathbb{R}^{\infty \times \infty}$ can be defined as infinite matrix and each element can be represented as $d_{i,j} \in \mathcal{D}$ and $r_{i,j} \in \mathbb{R}^{\infty \times \infty}$, $i, j \to \infty$, respectively. $\mathcal{D}$ can converge, if and only if $d_{i,j}^k \cdot r_{i,j}$ should be $\mathcal{O}(\frac{1}{n^p})$.

$$D = \begin{bmatrix} d_1 \\ d_2 \\ \vdots \\ d_{n-1} \\ \vdots \end{bmatrix} \tag{C.21}$$

In Eq. (C.21), operator $D$ denotes an infinite dimensionality operator.

$$X = \begin{bmatrix} r_1 \\ r_2 \\ \vdots \\ r_{n-1} \\ \vdots \end{bmatrix} \tag{C.22}$$

In Eq. (C.21), without generality, input $X$ denotes an infinite dimensionality matrix.

Then, given the operator $D$ applied on input matrix $X$ as:

$$D \otimes X = \begin{bmatrix} d_1 r_1 \\ d_2 r_2 \\ \vdots \\ d_{n-1} r_{n-1} \\ \vdots \end{bmatrix} \tag{C.23}$$

Obviously, due to the inequality of norms in the infinite dimensionality space, we examine the $\ell_2$ norm as an example:

$$\|D \otimes X\|_2 = \sum_{i=1}^{\infty} (d_i r_i)^2 \tag{C.24}$$

Easily, we can conclude:

$$\|D^k \otimes X\|_2 = \sqrt{\sum_{i=1}^{\infty} (d_i^k r_i)^2} < \infty \Leftrightarrow \lim_{k \to \infty} (d_i^k r_i)^2 = \frac{1}{n^p} \; p > 1 \tag{C.24}$$

Therefore, we have proved the DEMAND converges in an infinite dimensionality space, if and only if each element of $D \cdot X$ as $\frac{1}{n^p}$, and $p > 1, n \in \mathbb{R}$.

# Appendix D

*Lemma 4.1* (**Convergence of Alternative Optimization of Real Function**) For a series of real function as $\{f_{i,j}\}_{i,j=1}^{\infty}$. If we have: $\lim_{i\to\infty} f_{i,j}(x) \to h_{M,j}, a.e.\, x \in [a,b]$ and $\lim_{j\to\infty} h_{M,j} \to g_{M,N}, a.e.\, x \in [a,b]$. Then, $\exists \lim_{k\to\infty} f_{i_k,j_k} \to g_{M,N}\, a.e.\, x \in [a,b]$ holds.

*Proof*: If considering the uniform convergence of $\{f_{i,j}\}_{i,j=1}^{\infty}$, since $\lim_{i\to\infty} f_{i,j} \to h_{M,j}, a.e.\, x \in [0,1]$ and $\lim_{j\to\infty} h_{M,j} \to g_{M,N}, a.e.\, x \in [0,1]$, according to Riez Theorem, $\exists \{f_{i_k,j_k}\}_{k=1}^{\infty}$.

$$|f_{i_k,j_k} - h_j| < \frac{\varepsilon}{2} \quad (D.1)$$

And we have:
$$|h_j - g| < \frac{\varepsilon}{2} \quad (D.2)$$

Then, we have:
$$|f_{i_k,j_k} - h_j| + |h_j - g| = |f_{i_k,j_k} - g| < \varepsilon \quad (D.3)$$

*Theorem 4.1* (**Alternative Convergence of DEMAND**) If considering the continuous operator applied on finite dimensional space, the series of operators, $\{\mathcal{F}_{i,j}\}_{i,j=1}^{\infty}$, $\{\mathcal{H}_j\}_{j=1}^{\infty}$. And $\mathcal{F}_{i,j}: \mathbb{R}^{M\times N} \to \mathbb{R}^{M\times N}$. $\mathcal{H}_j: \mathbb{R}^{M\times N} \to \mathbb{R}^{M\times N}$. If we have: $\lim_{i\to\infty} \mathcal{F}_{i,j} \to \mathcal{H}_{M,j}$ and $\lim_{j\to\infty} \mathcal{H}_{M,j} \to \mathcal{G}$. Then, $\exists \lim_{k\to\infty} \mathcal{F}_{i_k,j_k} \to \mathcal{G}$ holds.

*Proof*: According to Lemma 1.5, similarly, let constant $T < \infty$, we have:
$$\|\mathcal{F}_{i_k,j_k} - \mathcal{H}_j\| < \frac{\varepsilon}{2T} \quad (D.4)$$

And we have:
$$\|\mathcal{H}_j - \mathcal{G}\| < \frac{\varepsilon}{2T} \quad (D.5)$$

The following inequality holds:
$$\|\mathcal{F}_{i_k,j_k} - \mathcal{H}_j\| + \|\mathcal{H}_j - \mathcal{G}\| = \|\mathcal{F}_{i_k,j_k} - \mathcal{G}\| < \frac{\varepsilon}{T} \quad (D.6)$$

And we also have:
$$\|\mathcal{F}_{i_k,j_k} X - \mathcal{G}X\| \leq \|\mathcal{F}_{i_k,j_k} - \mathcal{G}\| \cdot \|X\| < T \cdot \frac{\varepsilon}{T} = \varepsilon \quad (D.7)$$

This equation indicates the operator can converge to a fixed point defined on Banach space, using alternative strategy and Banach Fixed Point Theorem (Rudin, 1973), if and only if $\lim_{k\to\infty} \|\mathcal{F}_{i_k,j_k}\| < 1$.

*Theorem 4.2* (**Explanation of More Components Detected via DEMAND Than ICA**) In general, DEMAND can detect more components from an input signal than Independent Component Analysis Method.

*Proof*: At first, we assume all component included in signal matrix as:
$$I \stackrel{\text{def}}{=} \bigcup_{i=1}^{M} \xi_i \subseteq \mathbb{R}^{T\times M} \quad (D.8)$$

A single component can be denoted as following:
$$\xi_i \stackrel{\text{def}}{=} [\xi_{1,i}, \xi_{2,i}, \cdots \xi_{T,i}] \tag{D.9}$$

And we assume that there is no any overlap in these components:
$$\forall i \neq j \ \xi_i \cap \xi_j = \emptyset \tag{D.10}$$

If we define ICA operator as below:
$$ICA \stackrel{\text{def}}{=} \mathcal{T} \colon \mathbb{R}^{T \times M} \to \mathbb{R}^{1 \times M} \tag{D.11}$$

Obviously, when ICA is applied on the input signal, it is easy to conclude:
$$\mathcal{T} \cdot I = \mathcal{T}\left(\bigcup_{i=1}^{M} \xi_i\right) = [\xi_1, \xi_2, \cdots, \xi_M] \tag{D.12}$$

Similarly, as previous definition of DEMAND as operator $\mathcal{D}$, we can have:
$$\mathcal{D} \cdot I = D\left(\bigcup_{i=1}^{M} \xi_i\right) = [\xi_1, \xi_2, \cdots, \xi_M, (\xi_1, \xi_2), (\xi_1, \xi_3), \cdots, (\xi_1, \xi_3, \cdots, \xi_k), \cdots] \tag{D.13}$$

It is easy to calculate the number of components detected by ICA, due to independent constraint:
$$|\mathcal{T} \cdot I| = M \tag{D.14}$$

Nevertheless, we can conclude:
$$|\mathcal{D} \cdot I| = 2^M \tag{D.15}$$

Obviously, we also have:
$$M \ll 2^M \tag{D.16}$$

Inequality (D.16) demonstrates that the number of components identified by DEMAND should be more than ICAs.

# Appendix E

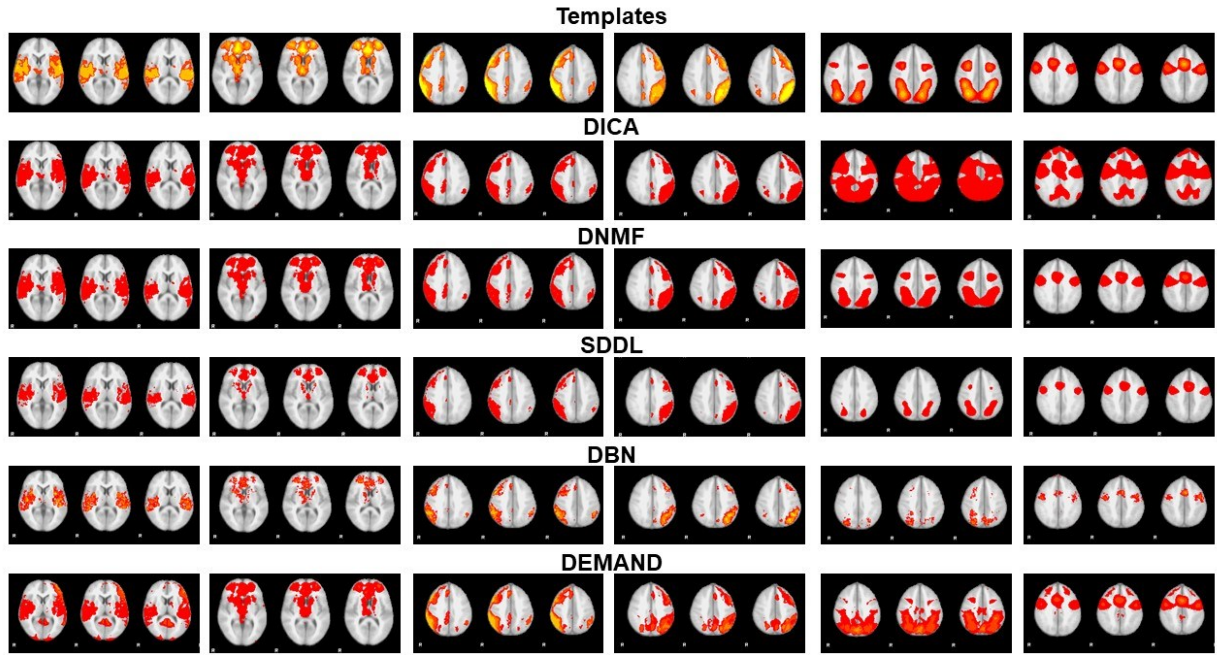

**Figure S2**. A qualitative comparison of another six reconstructed canonical FCs by DEMAND and other four peer algorithms.